# MARTI-4: new model of human brain, considering neocortex and basal ganglia – learns to play Atari game by reinforcement learning on a single CPU.

Igor Pivovarov[1][0000-0002-5701-8717] and Sergey Shumsky [1]

[1] Moscow Institute of Physics and Technology, Moscow, Russia

`igorpivovarov@yandex.ru`

**Abstract.** We present Deep Control – new ML architecture of cortico-striatal brain circuits, which use whole cortical column as a structural element, instead of a singe neuron. Based on this architecture, we present MARTI - new model of human brain, considering neocortex and basal ganglia. This model is designed to implement expedient behavior and is capable to learn and achieve goals in unknown environments. We introduce a novel *surprise feeling* mechanism, that significantly improves reinforcement learning process through inner rewards. We use OpenAI Gym environment to demonstrate MARTI learning on a single CPU just in several hours.

## 1 Introduction

In this work we introduce two new concepts. First is Deep Control Architecture - new hierarchical model of cortico-striatal brain circuits, which use a cortical column as a structural element, instead of a singe neuron. DCA is a hybrid vector-symbolic model, making native representations from high dimensional vector space to symbols and vice versa. Through this, DCA is very fast and compact way for real time learning, hierarchical analysis of environment, hierarchical planning and executing.

Second is MARTI – new ML model of human brain, built on Deep Control Architecture, implementing neocortex and basal ganglia. It runs ensemble of cortical columns simultaneously, orchestrated by basal ganglia, which is selecting the winner and inhibiting the rest of columns. Basal ganglia also maintains *surprise feeling*, which is a mechanism of implementation of inner rewards, allowing model to learn much faster. This multi-agent model is capable of learning by reinforcement learning to achieve goals in unknown environments.

To demonstrate  MARTI capabilities, we use OpenAI Gym Atari game Ping-Pong. We run both MARTI and Gym on a usual single CPU machine. Using this set-up MARTI robustly learns to play Ping-Pong game in several hours.

In this work we show the role of basal ganglia in a whole decision making process and conclude, that Deep Control Architecture is a new promising way of modeling human brain, especially where fast performance is needed with limited resources.

## 2 Background

Deep neural network is a low-level model of human neocortex, particularly visual cortex, which is perfectly designed for object detection/classification. However, DNN results in other domains, e.g. planning, decision making and appropriate behavior are far less impressive. Possibly, this is because behavior tasks are mostly implemented in other parts of human brain, besides neocortex.

Neocortex receives sensorimotor information, classify it and build «map of objects» and their relations. Positive feedback loops between thalamus and cortex supports long-time cortex activation, to allow synchronization between distant parts of brain. Basal ganglia, being the main keepers of values, can inhibit or disinhibit these positive feedback loops, being the main conductor of the cortex activity. Finally, cerebellum helps to maintain routine operations, adopting patterns, that were found previously by neocortex and basal ganglia. [1]

In this process neocortex plays important role, analyzing situation and predicting situation development, but it is basal ganglia, that plays key role in deciding on variants and implementation of most valuable variant. To implement behaivour tasks, one should propose a unified model of basal ganglia and neocortex.

## 3 Related work

Deep Control architecture proposed in this paper reflects biological mechanisms of the brain, namely the concept of hierarchical predictive coding of information in the neocortex [2, 3, 4, 5] Unlike other models of the neocortex[6, 7, 8], Deep Control integrates Hebbian learning in the cortex with reinforcement learning in basal ganglia, implementing so called super-learning architecture [9].

Learning hierarchies of policies is a long-standing problem in RL [10,11]. Namely [12] introduced the concept of options as closed-loop policies for taking action over a period of time, and [13] proposed option-critic architecture as an important step toward end-to-end hierarchical reinforcement learning. In these and similar works [14, 15] both goals and subgoals are defined in the same sensory-motor space. In our approach, each level operates in its own space using increasingly abstract representations to formulate higher levels plans.

## 4 Reinforcement learning environment

To evaluate behavioral tasks we use reinforcement learning approach. In current work, we used OpenAI Gym Atari games environment [16] and particularly Ping-Pong (PONG) game.

The Atari 2600 PONG game is one of the most complex games for reinforcement learning. Games can easily last 10,000 time steps (compared to 200-1000 in other domains); observations are also more complex, containing the two players' score and side walls. Pong paddle control is nonlinear: simple experimentation shows that fully predicting the player's paddle requires knowledge of the last 18 actions [17]. Finally, sparse rewards makes Pong quite complex game for RL.

We consider tasks in which an agent interacts with an environment $E$ (in this case the OpenAI Gym Atari emulator) in a sequence of actions, observations and rewards. At each time-step the agent selects an action $at$ from the set of available game actions, $A = \{0, \ldots K\}$. The action is passed to the emulator and modifies its internal state and the game score. Agent observes the $E$ state $st$ (it can be an image of current screen or any other representation of $E$ state). In addition it receives a reward $rt$ representing the change in game score. (In general the game score depends on the prior sequence of actions and observations and feedback about current action may only be received after many hundreds or thousands of time-steps have elapsed - this is so called *sparse rewards*.)

In this work our agent observes emulator state called RAM - bit memory state of Atari computer. As it was shown in [16], RAM state does not give some special advantages to agent and even controversial - it appears that screen image carries more structural information that is not easily extracted from the RAM bits, so neural networks usually learn better using screen image. But we use RAM representation here as a very rough model, based on idea, that behavioral centers of the human brain deal with preprocessed and good prepared data, not with raw images.

The goal of the agent is to interact with the emulator by selecting actions in a way that maximizes future rewards. Such model is not a perfect, but reasonable way to test abilities of ML model to learn and achieve goals in uncertain environments.

## 5 Deep Control Architecture (DCA)

Deep Control Architecture is a novel hierarchical model of human brain, including neocortex interaction with basal ganglia. First, we will discuss main ideas of DCA and then talk about current realization.

### 5.1 Main ideas of DCA

DCA represents a hierarchy of modules learning to jointly implement predictive behavior control with reinforcing signals coming from the dopamine system of the midbrain [18]. DCA comprise:

- a hierarchy of self-organizing maps of cortical modules, predicting activity of lower level cortical modules with primary sensory-motor modules at the lowest level;
- each hierarchical level corrects its predictions based on long-term predictions of the higher level and actual signals from the lower level;
- basal ganglia assess the usefulness of various patterns of cortical activity and select the winning pattern, implementing reinforcement learning

In general, DCA is based on the following premises:

**First,** DCA uses cortical *columns* rather than neurons as main functional units of neocortex. Thus, one have no more need to model each neuron. Considering various neurophysiological data, the basic structural elements of neocortex are cortical columns, each working with ~20-30 *symbols*.

**Second,** our conjecture is that several columns with local reciprocal connections form *hypercolumn*, capable of memorizing typical temporal patterns - *sequences of symbols*.

**Third** idea is about hierarchy. Hypercolumns alone cannot predict far enough into the future to solve complex tasks. But being organized in a hierarchy, higher levels operate at ever greater time scales, using sequences of lower level symbols as their input.

## 5.2    DCA structure

Based on these ideas, we introduce DCA as follows:
Cortical hypercolumn (CHC) is an autonomous module, working with vector data. CHC consists of two parts:

- Coder/Decoder – preprocessing high dimensional input vectors to discrete symbolic representation and back.
- Parser – processes symbolic data flow, finds patterns and regularity in data and predicts next symbols.

To create a new CHC, initial dataset of input vectors is needed. Then Coder/Decoder runs clusterization of this dataset (we use K-means clustering), mapping input vectors to K clusters. These clusters (or cluster numbers, if you like) become symbols for Parser. From this point, each new vector, received by CHC, is converted to symbol by Coder/Decoder and then processed by Parser.

## 5.3    Learning

Parser – processes symbolic data flow, finds patterns and regularity in data and predicts next symbol. For this purposes Parser has it's vocabulary $S$ with all the symbols and correlation table $C$, that keeps correlations between symbols. Each time Parser receives new symbol, $C$ is updated :

$$s_n \quad => \quad C\, s_{n-1},\, s_n = C\, s_{n-1},\, s_n + 1$$

If Parser has $m$ symbols in vocabulary and two symbols $s_{n-1}$ and $s_n$ are correlated more then defined threshold T, a new symbol (word) is formed and added to vocabulary:

$$if\, C\, s_{n-1},\, s_n > T \quad => \quad s_{m+1} = s_{n-1}s_n$$

Parser has predefined capacity of vocabulary size and word length, e.g. 1000 symbols and max word length = 3. Parser learns regularities in data and predicts next symbol.

### 5.4 Prediction

Prediction can be based on the correlation statistics – then we call it "situation prediction". In this case Parser predicts next symbol as follows:

$$s_{n+1} \implies \boldsymbol{max_i}(\boldsymbol{C}_{s_n,\ s_i})$$

Prediction can be based on value function. For this purposes Parser can keep reward table $\boldsymbol{R}$, that keeps rewards received after symbols. Each time Parser receives a non-zero reward, $\boldsymbol{R}$ is updated:

$$r_n \implies \boldsymbol{R}_{s_{n-1},\ s_n} = \boldsymbol{R}_{s_{n-1},\ s_n} + r_n$$

$$\boldsymbol{R}_{s_{n-2},\ s_{n-1}} = \boldsymbol{R}_{s_{n-2},\ s_{n-1}} + r_n * k$$

$$\dots$$

$$\boldsymbol{R}_{s_{n-m-1},\ s_{n-m}} = \boldsymbol{R}_{s_{n-m-1},\ s_{n-m}} + r_n * k^m,$$

where $m$ is predefined memory size

As a result, parser has working memory of rewards it received in particular situations. Based on $\boldsymbol{R}$ table, parser can predict *desired* next symbol with maximum expected reward (reward forecast):

$$s_{n+1} \implies \boldsymbol{max_i}(\boldsymbol{R}_{s_n, s_i})$$

Prediction is always a pair – next symbol $s_{n+1}$ and reward forecast of that next symbol $\boldsymbol{R}_{s_{n+1}}$. Which kind of prediction specific parser will use depends on architecture; we will discuss this in 6.

DCA use semantic coding to move to the next level of the hierarchy. State sequences of the lower level are mapped to the states of a higher level via clustering of probability vectors of their successors and predecessors. All hierarchical levels interact with each other, looking for a way to implement the abstract plan of a higher level, consistent with the newly received data from the lower one. The number of hierarchical levels increases with the increase in the amount of data collected when interacting with the environment. So does planning horizon, which makes the Deep Control architecture a good candidate for AGI.

## 6 MARTI

Furthermore, we present MARTI (Modular ARTificial Intelligence) – new model of human brain, built on Deep Control Architecture. In this work we present MARTI-4 prototype, implementing neocortex, basal ganglia and thalamus at a object level.

Neocortex is a set of hypercolumns, each of which acts as a autonomous agent, receives partial information from thalamus, converts it to it's own symbol representation and tries to predict next symbol. Basal ganglia striatum receive predictions from all columns and tries to figure out the most valuable action to continue with, selects the winner column and inhibits the rest. Thalamus serves as a main information hub, processing sensor and actuator information from outside, providing it to cortical hypercolumns and to basal ganglia and back. Thalamus also inhibits execution of hypercolumns, that do not have new input data.

MARTI-4 receives sensor data (environment state) $s_t$ and actuator data $a_t$ as input data, as well as current reward $r_t$.

## 6.1    First layer hypercolumns

At the initialization, thalamus uses random sampling to create $p$ subsets of size $m$ from the initial sensor data $s_t$. For each of this subsets thalamus creates $s$ cortical hypercolumns. Then, each time upon receiving new data, thalamus repeats this sampling to $p$ parts and feeds each part to corresponding column Coder.

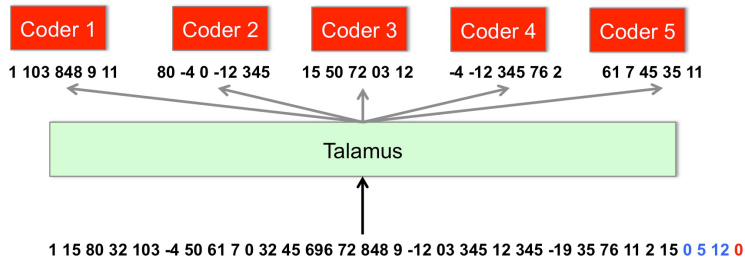

At first steps, there is no learning. At this stage Coder of each CHC is gathering data to create Parser. The condition of creating a Parser is that number of unique vectors received by this Coder exceeds given limit $v$. (Most of Coders never exceed this limit, because of different frequency of each of coordinates in initial vector). After limit $v$ is reached, Coder creates corresponding Parser as follows:

- Coder run clusterization of $v$ vectors, dividing vectors subset to $K$ clusters
- Each cluster receives a symbolic name – a letter in UTF-8, e.g. "A" to "Z"
- Parser object is created with this alphabet

From this step, each time Coder receives a new vector, it classifies this vector (based on it's clusterization) and gives Parser corresponding cluster symbolic name as an input.

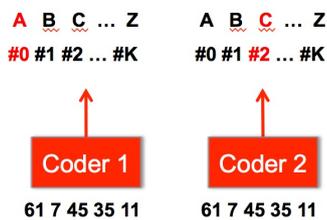

Parser of 1st layer in MARTI-4 is created with those restrictions:

- maximum word length = 1
- prediction type = situation (correlation based)

Parser task is to parse it's symbolic inputs and build a correlation table **C**, using which it can predict next symbol. This parser also has reward table **R**, but it is not used for predictions, it is used for calculating *surprise feeling*, which we will discuss later.

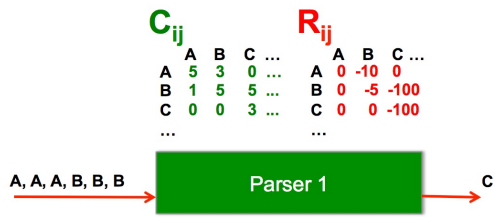

So, Parsers of 1st layer are very simple and fast, they do not build new symbols and work only with letters. They predict the most probable next symbol.

### 6.2    Action Coder/Decoder

After creating at least one Parser, thalamus creates special actuator Coder A for actuator data $a_t$ as follows:

- Coder A runs clusterization of actuator vectors subset $a_t$, dividing it to $K$ clusters
- Each cluster receives a symbolic name – a letter in UTF-8, e.g. "a" to "z"
- no Parser is created for this Coder A

From this step, each actuator vector is classified by Coder A (based on it's clusterization) and converted to corresponding cluster symbolic name - which represents current *action*.

### 6.3    Second layer hypercolumns

After creation of at least 3 hypercolumns of 1st layer, next layer is created as follows:

- Each 3 hypercolumns of 1st layer become a substrate to create hypercolumn of 2nd layer.
- Coder of 2nd layer hypercolumn combines symbols of 1st layer subcolumns with current action symbol to build symbol for it's Parser.

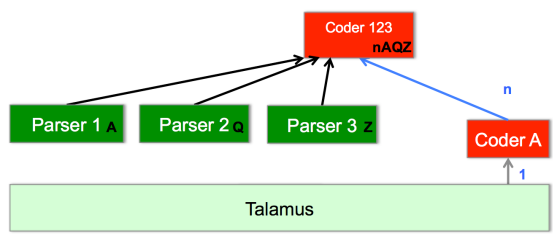

So, Parsers of 2<sup>nd</sup> layer works with symbols, combined from lower sensor symbols and action symbol, starting from action e.g. "nABC" or "dXYZ".They are created with restrictions:

- max word length = 4
- max vocabulary size = 5000
- prediction type = value (reward based)

Parsers of 2<sup>nd</sup> layer has reward table **R**, keeping summarized reward received after each symbol as was discussed in 5.2.

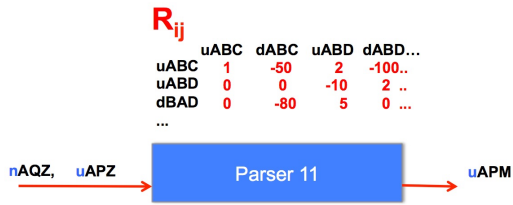

At each step each Parser predicts most valuable next symbol, which will maximize the future reward. Besides it, each Parser calculate it's «positive feeling» of all possible actions. It is calculated as overall *number* of positive reward symbols, beginning from this action:

$$F^v(a_i) = \Sigma_j \ I \mid \boldsymbol{R}(s_n, s_j) > 0 \qquad where \ \ s_j \mid s_j \in a_i$$

### 6.4 Basal ganglia

After all hypercolumns made their predictions, thalamus passes all the data to basal ganglia (striatum) to find the best prediction and, as a result, choose next action.

This is the most intriguing part of this paper, because most of usual RL approaches to choose next action does not work properly in this situation. We did a lot of experiments to find out working solution.

Usually, our intuition says, that in reinforcement learning approach model should take next action, which has maximum value function (or maximum future reward). In this case, that could mean choosing hypercolumn with maximum predicted reward. But suprisingly, at every moment we can find a hypercolumn giving a very high predicted reward combined with a wrong action. No separate hypercolumn can give a good prediction, because all of them have only partial sensor information. This is like CHC-1 "sees" only X coordinate of an object and CHC-2 "sees" only Y coordinate. *Their* predictions are always *biased* with *their* information.

That's why, to obtain better prediction, an ensemble of hypercolumns is needed. And the task of basal ganglia is to choose most promising way to increase future rewards.

In MARTI-4 basal ganglia striatum works as follows:

- For each action $a_i$ calculate "basal positive feeling" as number of CHC, that has $F^v(a_i) > 0$

$$F^b(a_i) = \Sigma_j \; 1 \; | \; F^c_j(a_i) > 0$$

- Choose next action $a_i$, which has maximum $F^b(a_i)$
- Select as a winner CHC, that predicted next symbol with this action $a_i$, which has maximum predicted future reward.

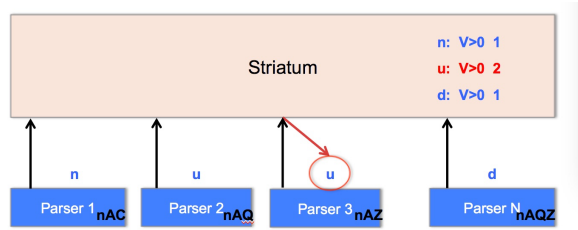

## 6.5 Surprise feeling and inner rewards

Another important task of basal ganglia is maintaining a *surprise feeling*, which helps the model understand what was done right. In reinforcement learning environment *sparse rewards* are big issue, because reward can be received after many hundreds or thousands of steps have elapsed. In this case, it will be nice to have any way of understanding, that something has been done properly right now, without waiting too long for a distant (and rare) reward.

To do it, basal ganglia analyze the state of each hypercolumn just after it received new data but before it made any predictions. Each parser compares new data with previous prediction it made. Prediction is always a pair – next symbol $s_{n+1}$ and reward forecast of that next symbol $\boldsymbol{R}s_{n+1}$. Similarly, received data also constructs a pair – symbol received $s_t$ and reward forecast of this symbol $\boldsymbol{R}s_t$. And if $s_t \mathrel{!=} s_{n+1}$ then reward forecast may have changed.

Hypercolumn *surprise feeling* can be defined as *unexpected improving of reward forecast:*

$$S^c(s_t) > 0 \; | \; \boldsymbol{R}s_t >> \boldsymbol{R}s_{n+1}$$

Note, that, especially for parsers of $1^{st}$ layer, usually parser receives (statistically) expected data and usually has expected deterioration of the reward forecast. But single surprise of single hypercolumn is not enough to be sure, that overall forecast became better. Basal ganglia observes all hypercolumns and calculate "basal surprise feeling" based on simultaneous surprises of different columns or sequential surprises of single column. When this overall surprise feeling becomes greater than given threshold $S^t$ a one time inner reward is given to all hypercolumns:

$$S^b(s_t) = \Sigma_j \; 1 \; | \; S^c_j(s_t) > 0$$

$$r_t = 1 \; | \; S^b(s_t) > S^t$$

This mechanism of *surprise feeling* allows model to learn much faster through implementation of inner rewards in addition to usual environment rewards.

### 6.6 Whole cycle of analysing/predicting

Finally, let's have an overview of the whole model work.

Each step thalamus receives sensor data $s_t$, actuator data $a_t$, current reward $r_t$. It samples $s_t$ to $p$ parts and feed each part to CHC of the 1st layer.

Each CHC of 1st layer encodes it's input vector to it's own symbols and processes symbol parsing taking into account "column surprise feeling" based on match between predicted and received symbols. After parsing, CHC makes prediction of next symbol and reward forecast.

Each CHC of 2nd layer receives symbols from 1st layers as an input and encodes them to own symbol representation. Then it processes symbol parsing taking into account "column surprise feeling". After parsing, CHC calculates next symbol prediction and "column positive feeling" for every potential action.

Basal ganglia striatum observe "column surprise feelings" from all CHC and calculates "basal surprise feeling" $S^b(s_t)$. If it is greater then a threshold $S^t$ then one time inner reward is given to all CHC.

Basal ganglia striatum receive predictions from CHC of 2nd layer and calculates "basal positive feeling" for each potential action. Then it chooses the winner, that has maximum future reward for action with maximum "basal positive feeling".

Coder A decodes predicted action back to actuator terms $a_t$ and it is returned back by thalamus to environment as a next action.

### 6.7 World model and prediction horizon

In terms of reinforcement learning, MARTI-4 is a model-based agent, because it builds it's own "world model" and use the information about last steps to understand it's position in this world model. However, prediction horizon in MARTI-4 is usually 2-3 steps forward, because it has only 2 hierarchical layers and the prediction horizon in DCA depends only on hierarchy levels. This will be the subject of future works.

## 7 Experiments

Experimental setup was standalone single CPU test machine (AMD A8-9600 10 compute cores 16Gb RAM). We used Ubuntu 18.04, python 3.6.9, OpenAI Gym library, Java OpenJDK 11.0.13 installed. Marti-4 is written in Java and is using simple TCP/IP socket interface to receive and send data. To connect it with OpenAI Gym, we use additional python script, which receive data from Gym and send it to Marti via socket. When Marti have processed sensor data, it sends back actuator data to script, which sends it to Gym.

In previous works [17, 19], to set an experiment to test a model performance in PONG game, researches choose to run it up to 18,000 game steps or up to score 21. We found this setup not the best way to reveal model performance, because of specif-ic Pong game nature: if model learns how to beat back the ball in some situations, that usually will not lead to win the game or even to win a single play in a game, because other side (Gym) will beat back the ball in most cases and finally win the game play. In other words, even when model steadily learns how to play, it still loses when play up to 21 score or 18,000 steps.

So, in our setup we choose to run the game up to 500 steps. If the model is not able to beat back at all, overall score is usually around 0:15. But when model steadily learns to play and beats back more and more, the plays become longer and longer and score looks like 1:1 or 3:4 or something like that.

To perform an experiment, both MARTI and Gym are executed on a test machine. CPU load is about 40% for Gym and 60% for MARTI. MARTI size in memory is ~8Gb. Using this setup MARTI robustly learns to play Ping-Pong game in 500-700 game plays (3-4 hours).

Typical experimental run is shown in Fig. 1 One can see, that starting from typical score 0:15 with average goal difference is -15, MARTI makes quick progress and reaches typical scores 2:2, 1:0 with average goal difference -1 in 500-700 games.

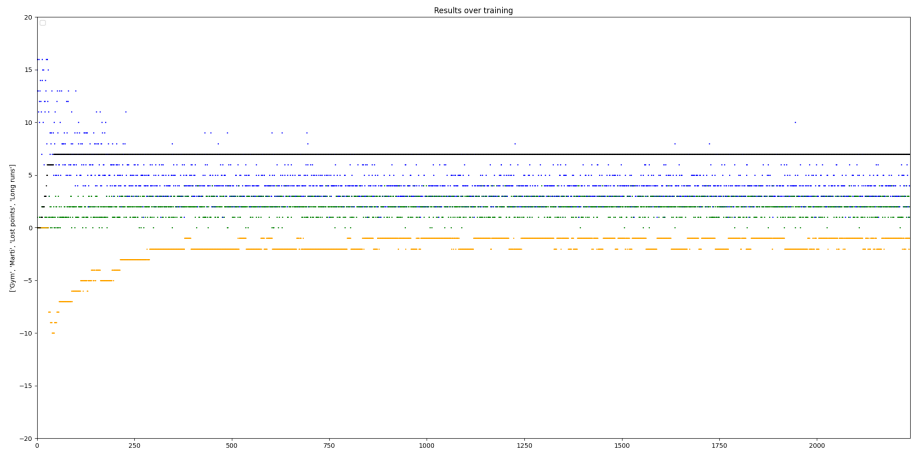

**Fig. 1.** *On X-axe there is number of games played, each 500 steps. On Y-axe there are points: blue dots are Gym goals in the game, green dots are Marti goals in the game, orange points are goal difference between Gym and Marti calculated as average last at 30 games..*

For purposes of comparison with previous works, we also preformed evaluation of model performance as in [19]. An episode starts on the frame that follows the reset command, and terminates when the end-of-game condition is detected or after 5 minutes of real-time play (18,000 frames), whichever comes first. A trial consists of 500 training episodes, followed by 500 evaluation episodes. Agent's performance was measured as the average score achieved during the evaluation episodes across 3 sequential trials. This setup is consistent to setups used in [17, 19] with the only difference, that MARTI-4 does not show significant improvement after 500 training epi-sodes, so training episodes were lowered to 500. Table 1 shows MARTI performance, compared to previous works in this setup gives a summary of all heading levels.

**Table 1.** Performance of different algorithms on PONG game.

| ALGORITHM | PONG |
|-----------|------|
| Random [17] | -20.9 |
| Sarsa [17] | -19 |
| MARTI-4 | -15,8 |
| Human [19] | -3 |
| UCT [17] | 21 |

Since MARTI-4 has only 2 layers, it can hardly been compared with deep networks like DQN. However, even this small model shows comparable results with models like Sarsa.

## 8    Discussion

Current model has modest results and never get to score *21:0*. This is because current prototype has only 2 hierarchical layers of hypercolumns and full power of DCA will be obtained, when there will be much more layers of hypercolumns, hierarchically organized. So, current work can be considered only as a testbed for this way of modeling. However, we demonstrate that even this simple model is capable to learn in unknown environment and show quick progress.

One of the reasons, DCA architecture is very fast is because model is building "on the fly" from zero, model size and hierarchy depends only on the amount and variety of input data. This is in contrast to deep neural networks, that are build initially huge and one have a need to run calculations forward and back through all this billions on neurons.

DCA perfectly suited to work with data preprocessed with DNN. Next thing to do is to make model input not a RAM state, but raw screen images, preprocessed with CNN. This will be more similar to real process, which take place in human brain. Last but not least final technical issue is that neither OpenAI Gym Atari emulator nor ALE Atari emulator are providing fully reliable and expected behavior of Atari game. Namely, in some cases (1 of ~50 games) some unexpected behavior of Atari emulator occurs, when the gameplay is already finished, new gameplay should start, but screen remains unchanged for some time and model continues to receive some environmental data which makes no sense. In some games (Atari Breakout for example) this can last for 30,000 steps and more. This makes the learning process significantly more complicated.

## 9    Conclusion

We showed, that Deep Control Architecture is a hybrid vector-symbolic ML architecture, making native representations from high dimensional vector space to symbols and back. Through this, DCA is very fast and compact way for real time learning, hierarchical analysis of environment, hierarchical planning and executing, especially where fast performance with low resources is needed.

We presented MARTI - novel ML model of human brain, implementing neocortex, basal ganglia and thalamus, capable to learn by reinforcement learning to achieve goals in unknown environments. We presented a novel *surprise feeling* mechanism, that significantly improves reinforcement learning process through inner rewards.

Through this work we also tried to show the role and potential mechanism of basal ganglia work in human brain in a whole decision making process.

### Disclaimer

Igor Pivovarov works part time in Moscow Institute of Physics and Technologies, Huawei, Skoltech, Bauman University and IP Laboratories. Sergey Shumsky works part time in Moscow Institute of Physics and Technologies and Bauman University. However, the whole scope of current work was made by authors solely in free time without any support or participation of any entities.